%% file: main.tex
\documentclass[runningheads, envcountsame, a4paper]{llncs}
\usepackage[pdftex]{graphicx}
\usepackage[numbers]{natbib}
\usepackage{amsmath}
\usepackage{amsfonts}
\usepackage{placeins}
\usepackage{subfig}
\usepackage[export]{adjustbox}[2011/08/13] 
\usepackage[misc]{ifsym}
\usepackage{microtype}

\newcommand{\fullpaper}[1]{#1}
\newcommand{\ecml}[1]{}

\fullpaper{
\usepackage{hyperref}
\usepackage{color}

}

\ecml{
\usepackage[hyphens]{url}
}

\DeclareMathOperator*{\argmin}{argmin}
\DeclareMathOperator*{\argmax}{argmax}

\newcommand{\x}{\mathbf{x}}
\newcommand{\y}{\mathbf{y}}
\newcommand{\s}{\mathbf{s}}
\newcommand{\e}{\mathbf{e}}

\newcommand{\uci}{4}

\begin{document}
\title{Beyond the Selected Completely At Random Assumption for Learning from Positive and Unlabeled Data}
\titlerunning{Beyond the SCAR Assumption for PU Learning}
%
\author{Jessa~Bekker\orcidID{0000-0003-1928-7374} (\Letter) \and
Pieter~Robberechts\orcidID{0000-0002-3734-0047} \and
Jesse~Davis\orcidID{0000-0002-3748-9263}}
    \tocauthor{Jessa~Bekker, Pieter~Robberechts, and Jesse~Davis}
\toctitle{Beyond the Selected Completely At Random Assumption for Learning from Positive and Unlabeled Data}
\authorrunning{J. Bekker et al.}
%
\institute{Dept of Computer Science, KU Leuven, Belgium\\
\email{{first.lastname}@cs.kuleuven.be}}
\maketitle              
\begin{abstract}
	Most positive and unlabeled data is subject to selection biases. The labeled examples can, for example, be selected from the positive set because they are easier to obtain or more obviously positive. This paper investigates how learning can be enabled in this setting. We propose and theoretically analyze an empirical-risk-based method for incorporating the labeling mechanism. Additionally, we investigate under which assumptions learning is possible when the labeling mechanism is not fully understood and propose a practical method to enable this. Our empirical analysis supports the theoretical results and shows that taking into account the possibility of a selection bias, even when the labeling mechanism is unknown, improves the trained classifiers.
	
\end{abstract}
\section{Introduction}
Positive and unlabeled learning focuses on the setting where the training data contains some labeled positive examples and unlabeled examples, which could belong to either the positive or negative class. This contrasts to supervised learning, where a learner has a fully labeled training set and to semi-supervised learning, where a learner (usually) observes some labeled examples from each class. Positive and unlabeled (PU) data naturally arises in many applications. Electronic medical records (EMR) list diseases that a patient has been diagnosed with, however, many diseases are undiagnosed. Therefore, the absence of a diagnosis does not imply that a patient does not have the disease. Similarly, automatically constructed knowledge bases (KBs) are incomplete, and hence any absent tuple may be either true (i.e., belong in the knowledge base) or false~\citep{zupanc2018estimating}.

\citet{elkan2008learning} formalized one of the most commonly made assumptions in PU learning: the observed positive examples were selected completely at random from the set of all positive examples. This assumption means that the probability of observing the label of a positive example is constant (i.e., the same for all positive examples), which facilitates and simplifies both theoretical analysis and algorithmic design. This setting has been extensively explored in the literature~\cite{denis1998pac,lee2003learning,liu2003building,denis2005learning,mordelet2014bagging,claesen2015robust,hsieh2015pu,Chang2016PositiveUnlabeledLI,Blockeel2017PUlearningDC,Kiryo2017PositiveUnlabeledLW,Northcutt2017LearningWC,Hou2018GenerativeAP}. 

Unfortunately, the ``selected completely at random'' assumption is often violated in real-world data sets. For example, a patient's EMR will only contain a diagnosis if she visits a doctor, which will be influenced by factors such as the severity of the symptoms and her socio-economic status. 
The problem of biases in the observed labels has been recognized in the recommender systems and retrieval  literature~\cite{Schnabel2016RecommendationsAT,marlin:recsys09,joachims:wsdm17}. However, these works differ from PU learning in that the labels for some examples from each ``class'' are observed. Still, within the context of PU learning, there has been little (or no) work that focuses on coping with biases in the observed positive labels. 

The contribution of this paper is to take a step towards filling that gap by proposing and analyzing a new, less restrictive assumption for PU learning: the {\it Selected At Random (SAR)} assumption. Instead of assuming a constant probability for all positive examples to be labeled, it assumes that the probability is a function of a subset of an example's attributes. To help analyze this new setting, we leverage the idea of a propensity score, which is a term originating from the causal inference literature~\citep{imbens2015causal}. Intuitively, the propensity score can be thought of as an instance-specific probability that an example was selected to be labeled. We show theoretically how using propensity scores in a SAR setting provides benefits. Then, we discuss a practical approach for learning the propensity scores from the data and using them to learn a model. Empirically, we show that for SAR PU data, our approach results in improved performance over making the standard selected completely at random assumption. 






\section{Preliminaries}
PU learning entails learning a binary classifier only given access to positive examples and unlabeled data. This paper considers the single-training set scenario, where the data can be viewed as a set of triples $(x,y,s)$ with $x$ a vector of the attributes, $y$ the class and $s$ a binary variable representing whether the tuple was selected to be labeled. While $y$ is always missing, information about it can be derived from the value of $s$. If $s=1$, then the example belongs to the positive class as $\Pr(y=1|s=1)=1$. If $s=0$, the instance can belong to either class.

In PU learning, it is common to make the \emph{Selected Completely at Random (SCAR)} assumption, which assumes that the observed positive examples are a random subset of the complete set of positive examples. Selecting a positive example is therefore independent of the example's attributes $\Pr(s=1|y=1,x)=\Pr(s=1|y=1)$. The probability for selecting a positive example to be labeled is known as the label frequency $c=\Pr(s=1|y=1)$. A neat advantage of the SCAR assumption is that, using the label frequency, a model that predicts the probability of an example being labeled can be transformed to the classifier:
$\Pr(y=1|x)=\Pr(s=1|x)/c.$

Knowing the label frequency is equivalent to knowing the class prior $\alpha=\Pr(y=1)$ as one can be derived from the other: $c=\Pr(s=1)/\alpha$. Under the SCAR assumption, PU learning can therefore be reduced to estimating the class prior or label frequency and training a model to predict the observed labels.

Estimating the label frequency is an ill-defined problem because it is not identifiable: the absence of a label can be explained by either a small prior probability for the positive class or a low label frequency \citep{scott2015aro}. For the class prior to be identifiable,  additional assumptions are necessary.
Different assumptions have been proposed, but they are all based on attributing as many missing classes as possible to a lower label frequency as opposed to a lower positive class probability. The following assumptions are listed from strongest to strictly weaker. The strongest assumption is that the classes are non-overlapping, which makes the class prior and the labeled distribution match the unlabeled one as closely as possible~\cite{elkan2008learning,plessis2014classpe}. Others make the assumption that there exists a positive subdomain of the instance space, but the classes can overlap elsewhere~\cite{scott2015aro,plessis2015classprioref,bekker2018aaai}. \citet{ramaswamy2016mixturepe} assumes that a function exists which only selects positive instances. Finally, the irreducibility assumption states that the negative distribution cannot be a mixture that contains the positive distribution~\cite{Blanchard2010SemiSupervisedND,jain2016estimatingtc}.



\section{Labeling Mechanisms for PU Learning}
\label{sec:sar}
The labeling mechanism determines which positive examples are labeled. To date, PU learning has largely focused on the SCAR setting. However, labels are not missing completely at random in most real-world problems.  For example, facts in automatically constructed KBs are biased in several ways. One, they are learned from Web data, and only certain types of information appear on the Web (e.g., there is more text about high-level professional sports teams than low-level ones).  Two, the algorithms that extract information from the Web employ heuristics to ensure that only information that is likely to be accurate (e.g., redundancy) is included in the KB. Similarly, biases arise when people decide to like items online, bookmark web pages, or subscribe to mail lists. Therefore, we believe it is important to consider and study other labeling mechanisms.

When \citet{elkan2008learning} first formalized the SCAR assumption, they noted the similarity of the PU setting to the general problem of learning in the presence of missing data. Specifically, they noted that the SCAR assumption is somewhat analogous with the missing data mechanism called \emph{Missing Completely At Random (MCAR)} \citep{rubin1976inference}. Apart from MCAR, the two other classes of missing data mechanisms are \emph{Missing At Random (MAR)} and \emph{Missing Not At Random (MNAR)}. To complete this analogy, we propose the following corresponding classes of PU labeling mechanisms:
\begin{description}
	\item[SCAR] \emph{Selected Completely At Random}: The labeling mechanism does not depend on the attributes of the example, nor on the probability of the example being positive: each positive example has the same probability to be labeled.
	\item[SAR]  \emph{Selected At Random}: The labeling mechanism depends on the values of the attributes of the example, but given the attribute values it does not depend on the probability of the example being positive.
	\item[SNAR] \emph{Selected Not At Random}: All other cases: The labeling mechanism depends on the real probability of this example being positive, even given the attribute values.
\end{description}

There is one very important difference between PU labeling mechanisms and missingness mechanisms in that the labeling always depends on the class value: only positive examples can be selected to be labeled. According to the missingness taxonomy, all PU labeling mechanisms are therefore MNAR. SNAR is a peculiar class because it depends on the real class probability, while the class needs to be positive by definition. The class probability refers to the probability of an identical instance to this one being positive. Consider, for example, the problem of classifying pages as interesting. If a page is moderately interesting to you, some days you might like it while other days you might not. The labeling mechanism in this case could depend on how much you like them and therefore on the instance's class probability.

\section{Learning with SAR Labeling Mechanisms}


In this paper, we focus on SAR labeling mechanisms, where the key question is how we can enable learning from SAR PU data.
Our key insight is that the labeling mechanism is also related to the notion of a propensity score from causal inference~\citep{imbens2015causal}. 
In causal inference, the propensity score is the probability that an instance is assigned to the treatment or control group. This probability is instance-specific and based on a set of the instance's attributes. 
We use an analogous idea and define the propensity score as the labeling probability for positive examples:
\begin{definition}[Propensity Score] The propensity score for  $x$, denoted $e(x)$, is the label assignment probability for positive instances with attributes $x$,
	\[
	e(x) = \Pr(s=1|y=1,x).
	\]
\end{definition}
\noindent A crucial difference with the propensity score from causal inference is that our score is conditioned on the class being positive. 


We incorporate the propensity score when learning in a PU setting by using the propensity scores to reweight the data. Our scheme generalizes an approach taken for SCAR data~\cite{steinberg1992estimating,Plessis2015ConvexFF,Kiryo2017PositiveUnlabeledLW} to the SAR setting. In causal inference, inverse-propensity-scoring is a standard method where the examples are weighted with the inverse of their propensity score \citep{little2002statistical,imbens2015causal,Schnabel2016RecommendationsAT}. This cannot be applied when working with positive and unlabeled data, because we have zero probability for labeling negative examples. But we can do a different kind of weighting. The insight is that for each labeled example $(x_i,s=1)$ that has a propensity score $e_i$, there are expected to be $\frac{1}{e_i}$ positive examples, of which $\frac{1}{e_i}-1$ did not get selected to be labeled. This insight can be used in algorithms that use counts, to estimate the correct count from the observed positives and their respective propensity scores.  In general, this can be formulated as learning with negative weights: every labeled example gets a weight $\frac{1}{e_i}$ and for every labeled example a negative example is added to the dataset that gets a negative weight $1-\frac{1}{e_i}$.

We now provide a theoretical analysis of the propensity-weighted method, to characterize its appropriateness. We consider two cases: (1) when we know the true propensity scores and (2) when we must estimate them from data. All the proofs are deferred to the appendix\ecml{ (\url{https://dtai.cs.kuleuven.be/software/sar})}.

\subsection{Case 1: True Propensity Scores Known}
Standard evaluation measures, such as Mean Absolute Error (MAE), Mean Square Error (MSE) and log loss, can be formulated as follows:
\begin{align*}
	R(\hat\y|\y) = \frac{1}{n}\sum_{i=1}^n y_i\delta_1(\hat y_i) + (1- y_i)\delta_0(\hat y_i),\footnotemark
\end{align*}
\footnotetext{We assume that $0 < \hat y < 1$}
where $\y$ and $\hat\y$ are vectors of size $n$ containing, respectively, the true labels and predicted labels. The function $\delta_y(\hat y)$ represents the cost for predicting $\hat y$ when the class is $y$, for example:
\begin{align*}
	\text{MAE}&: \delta_y(\hat y) = |y - \hat y|,\\
	\text{MSE}&: \delta_y(\hat y) = (y-\hat y)^2,\\
	\text{Log Loss}&: \delta_1(\hat y) = -\ln \hat y~,~\delta_0(\hat y) = -\ln(1-\hat y).
\end{align*}

We can formulate propensity-weighted variants of these cost functions as: 

\begin{definition}[Propensity-Weighted Estimator]
	Given the propensity scores $\e$ and PU labels $\s$, the propensity weighted estimator of $R(\hat \y|\y)$ is
	\begin{align*}
		\hat R(\hat \y|\e,\s) = \frac{1}{n}\sum_{i=1}^n &s_i\left(\frac{1}{e_i}\delta_1(\hat y_i) + (1-\frac{1}{e_i})\delta_0(\hat y_i)\right) + (1-s_i)\delta_0(\hat y_i) ,
	\end{align*}
	where $\y$ and $\hat\y$ are vectors of size $n$ containing, respectively, the true labels and predicted labels. The function $\delta_y(\hat y)$ represents the cost for predicting $\hat y$ when the class is $y$.
\end{definition}

This estimator is unbiased:
\begin{align*}
	&\mathbb{E}[\hat R(\hat \y|\e,\s)])\\
	&\qquad = \frac{1}{n}\sum_{i=1}^n y_i e_i\left(\frac{1}{e_i}\delta_1(\hat y_i) + (1-\frac{1}{e_i})\delta_0(\hat y_i)\right) + (1-y_i e_i)\delta_0(\hat y_i)\\
	&\qquad = \frac{1}{n}\sum_{i=1}^n y_i \delta_1(\hat y_i) + (1-y_i)\delta_0(\hat y_i) \\
	&\qquad = R(\hat \y|\y).
\end{align*}

To characterize how much the estimator can vary from the expected value, we provide the following bound:

\begin{proposition}[Propensity-Weighted Estimator Bound]
	\label{the:Rbound}
	For any predicted classes $\hat \y$ and real classes $\y$ of size $n$, with probability $1-\eta$, the propensity-weighted estimator $\hat R(\hat \y|\e,\s)$ does not differ from the true evaluation measure $R(\hat \y|\y)$ more than
	\begin{align*}
		|\hat R(\hat \y|\e,\s)-R(\hat \y|\y)|\leq \sqrt{\frac{\delta_\text{max}^2\ln\frac{2}{\eta}}{2n}},
	\end{align*}
	with $\delta_\text{max}$ the maximum absolute value of cost function $\delta_y$.
\end{proposition}

The propensity-weighted estimator can be used as the risk for Empirical Risk Minimization (ERM), which searches for a model in the hypothesis space $\mathcal{H}$ by minimizing the risk: 
\begin{align*}
	\hat \y_{\hat R} = \argmin_{\hat \y_\in \mathcal{H}} \hat R(\hat \y|\e,\s).
\end{align*}
\noindent The following proposition characterizes how much the estimated risk for hypothesis $\hat \y_{\hat R}$ can deviate from its true risk.


\begin{proposition}[Propensity-Weighted ERM Generalization Error Bound]
	\label{the:genBound}
	For a finite hypothesis space $\mathcal{H}$, the difference between the propensity-weighted risk of the empirical risk minimizer $\hat \y_{\hat R}$  and its true risk is bounded, with probability $1-\eta$, by:	
	\begin{align*}
		R(\hat \y_{\hat R}|\y) \leq \hat R(\hat \y_{\hat R}|\e,\s) + \sqrt{\frac{\delta_\text{max}^2\ln\frac{|\mathcal{H}|}{\eta}}{2n}}.
	\end{align*}
\end{proposition}

\subsection{Case 2: Propensity Scores Estimated from Data}

Often the exact propensity score is unknown, but we have an estimate $\hat e$ of it. In this case, the bias of the propensity-weighted estimator is:

\begin{proposition}[Propensity-Weighted Estimator Bias]
	\label{the:Rbias}

	\begin{align*}
		\text{bias}(\hat R(\hat \y|\hat \e,\s)) = \frac{1}{n}\sum_{i=1}^n y_i(1-\frac{e_i}{\hat e_i})\left(\delta_1(\hat y_i) -\delta_0(\hat y_i)\right).
	\end{align*}
\end{proposition}
From the bias, it follows that the propensity scores only need to be accurate for positive examples. An incorrect propensity score has a larger impact when the predicted classes have more extreme values (i.e., tend towards zero or one). Underestimated propensity scores are expected to result in a model with a higher bias. Lower propensity scores result in learning models that estimate the positive class to be more prevalent than it is, which results in a larger $\left(\delta_1(\hat y_i) -\delta_0(\hat y_i)\right)$ for positive examples.


\subsubsection{Side Note on Sub-Optimality of Expected Risk}
Another method that one might be inclined to use when incorporating the propensity score is to minimize the expected risk\footnote{Derivation in appendix\ecml{, available on \url{https://dtai.cs.kuleuven.be/software/sar}}}:

\begin{align*}
	\hat{R}_\text{exp}(\hat\y|\e,\s)&=\mathbb{E}_{\y|\e,\s,\hat \y}\left[R(\hat\y|\y)\right]\\
	&=\frac{1}{n}  \sum_{i=1}^n \left(s_i+(1-s_i)\frac{\hat y_i(1-e_i)}{1-\hat y_ie_i}\right)\delta_1(\hat y_i)+  (1-s_i)\frac{1-\hat y_i}{1-\hat y_ie_i}\delta_0(\hat y_i).
\end{align*}

However, the expected risk is not an unbiased estimator of the true risk and as a result, $\hat\y_{\hat R_\text{exp}}=\text{argmin}_{\hat\y\in \mathcal{H}}
\hat R_\text{exp}(\hat\y|\e,\s)$ is not expected to be the best hypothesis. In fact, the hypothesis of always predicting the positive class $\forall_i : \hat y_i=1$ always has an expected risk $ \hat R_\text{exp}(\hat\y|\e,\s)=0$. 

\section{Learning under the SAR Assumption}
If the propensity scores for all examples are known (i.e., the exact labeling mechanism is known), they can be directly incorporated into the learning algorithm. However, it is more likely that they are unknown. Therefore, this section investigates how to permit learning in the SAR setting when the exact propensity scores are unknown. We discuss two such settings. The first is interesting from a theoretical perspective and the second from a practical perspective. 


\subsection{Reducing SAR to SCAR}

Learning the propensity scores from positive and unlabeled data requires making additional assumptions: if any arbitrary instance can have any propensity score, then it is impossible to know if an instance did not get labeled because of a low propensity score or a low class probability. Therefore, the propensity score needs to depend on fewer attributes than the final classifier \citep{imbens2015causal}. A simple way to accomplish this is to assume that the propensity function only depends on  a subset of the attributes $x_e$ called the \emph{propensity attributes}: 

\begin{align*}
	\Pr(s=1|y=1,x) &= \Pr(s=1|y=1,x_e)\\
	e(x)&=e(x_e).
\end{align*}
Often, this a realistic assumption. It is trivially true if the labeling mechanism does not have access to all attributes (e.g., because some were collected later). It may also arise if a labeler cannot interpret some attributes (e.g., raw sensor values) or only uses the attributes that are known to be highly correlated with the class.

To see why this can be a sufficient assumption for learning in a SAR setting, consider the case where the propensity attributes $x_e$ have a finite number of configurations, which is true if these attributes are all discrete.  In this case, it is possible to partition the data into strata, with one stratum for each configuration of $x_e$. Within a stratum, the propensity score is a constant (i.e., all positive examples have the same propensity score) and can thus be determined using standard SCAR PU learning techniques. 
Note that, as previously discussed, the SCAR assumption alone is not enough to enable learning from PU data, and hence one of the additional assumptions~\cite{Blanchard2010SemiSupervisedND,scott2015aro,plessis2015classprioref,ramaswamy2016mixturepe,bekker2018aaai} must be made. 



Reducing SAR to SCAR is interesting because it demonstrates that learning in the SAR setting is possible. However, it is suboptimal in practice as it does not work if $x_e$ contains a continuous variable. Even for the discrete case, the number of configurations grows exponentially as the size of $x_e$ increases. Furthermore, information is lost by partitioning the data. Some smoothness of the classifier over the propensity attributes is expected, but this is not encouraged when learning different classifiers for each configuration. Similarly, the propensity score itself is expected to be a smooth function over the propensity variables.

\subsection{EM for Propensity Estimation}
The problems with reducing the SAR to the SCAR case motivate the need to jointly search for a classifier and lower dimensional propensity score function that best explain the observed data. This approach also offers the advantage that it relaxes the additional assumptions: if they hold in the majority of the propensity attributes' configurations, the models' smoothness helps to overcome potential issues arising in the configurations where the assumptions are violated. This subsection presents a simple expectation-maximization method for simultaneously training the classification and the propensity score model. It aims to maximize the expected log likelihood of the combination of models.

\paragraph{Expectation} Given the expected classification model $\hat f$ and propensity score model $\hat e$, the expected probability of the positive class $\hat y_i$ of  instance $x_i$ with label $s_i$ is:\footnote{All derivations for this section are in the appendix\ecml{, available on \url{https://dtai.cs.kuleuven.be/software/sar}}.}
\begin{align*}
	\hat y_i &= \Pr(y_i=1|s_i,x_i,\hat f,\hat e)\\
	&= s_i +  (1-s_i)\frac{\hat f(x_i)\left(1-\hat e(x_i)\right)}{1-\hat f(x_i)\hat e(x_i)}.
\end{align*} 
%
%
%
%
%

\paragraph{Maximization}
Given the expected probabilities of the positive class $\hat y_i$, the models $f$ and $e$ are trained to optimize the expected log likelihood:
\begin{align*}
	&\text{argmax}_{f,e} \sum_{i=1}^n \mathbb{E}_{y_i|x_i,s_i,\hat f,\hat e} \ln \Pr(x_i,s_i,y_i|f,e)\\
	&~= \text{argmax}_{f} \sum_{i=1}^n \big[\hat y_i \ln f(x_i) + (1-\hat y_i)\ln(1-f(x_i)) \big],\\
	&~ \quad \text{argmax}_{e}  \sum_{i=1}^n  \hat y_i \big[s_i\cdot\ln e(x_i)+(1-s_i)\cdot\ln (1-e(x_i))\big]\\
\end{align*}

From the maximization formula, it can be seen that to optimize the log likelihood, both models need to optimize the log loss of a weighted dataset. The classification model $f$ receives each example $i$ twice, once as positive, weighted by the expected probability of it being positive $\hat y_i$ and once as negative, weighted by the expected probability of it being negative $(1-\hat y_i)$. The propensity score model $e$ receives each example once, positive if the observed label is positive and negative otherwise, weighted by the expected probability of it being positive $\hat y_i$.

Because this approach minimizes log loss, it will work best if the classes are separable. If the classes are not separable, then the trained classification model is not expected to be the optimal one for the trained propensity model (see previous section). In that case, it is advisable to retrain the classifier with the obtained propensity score, using the propensity-weighted risk estimation method.

The classification model is initialized by fitting a balanced model which considers the unlabeled examples as negative.
The propensity score model is initialized by using the classification model to estimate the probabilities of each unlabeled example being positive. 

Classic EM converges when the log likelihood stops improving. However, the likelihood could stop improving before the propensity score model has converged.  Convergence is therefore formulated as convergence of both the log likelihood and the propensity model. We measure the change in the propensity score model by the average slope of the minimum square error line through the propensity score prediction of the last $n$ iterations.

\section{Empirical Evaluation}

This section illustrates empirically that the SAR assumption facilitates better learning from SAR PU data. We compare both SAR and SCAR methods, so that the gain of using an instance-dependent propensity score over a constant label frequency can be observed. More specifically, we address the following questions:
\begin{description}
	\item[Q1.] Does propensity score weighting (SAR) improve classification performance over assuming that data is SCAR and using class prior weighting?
	\item[Q2.] Can the propensity score function be recovered?
	\item[Q3.] Does the number of propensity attributes affect the performance?
\end{description}


\subsection{Data}
We use eight real-world datasets which cover a range of application domains such as text, images and tabular data. These datasets are summarized in Table~\ref{tab:data}. Since the 20 News Groups, Cover Type, Diabetes and Image Segmentation datasets are originally multi-class datasets, we first transformed them by either grouping or ignoring classes.
For 20 News Groups,\footnote{\url{http://archive.ics.uci.edu/ml/}} we distinguish between computer (pos) and recreational (neg) documents. After removing their headers, footers, quotes, and English stop words, the documents were transformed to 200 word occurrence attributes using the Scikit-Learn\footnote{\url{http://scikit-learn.org}} count vectorizer.
For Cover Type,\footnotemark[\uci] we distinguish the Lodgepole Pine (pos) from all other cover types (neg).
The Diabetes\footnotemark[\uci] data was preprocessed in a similar manner to \citet{strack2014impact}. Additionally, we dropped attributes with the same value in 95\% of the examples, and replaced uncommon attribute values by ``other''. The positive class is patients being readmitted within 30 days.
Image Segmentation\footnotemark[\uci] was used to distinguish between nature (sky, grass or foliage) and other scenes (brickface, cement, window, path).
Adult,\footnotemark[\uci] 
Breast Cancer,\footnotemark[\uci]
Mushroom,\footnotemark[\uci] and
Splice\footnote{Available on LIBSVM Data repository \url{https://www.csie.ntu.edu.tw/~cjlin/libsvmtools/datasets/}} were used as is. To enable using logistic regression, all the datasets were further preprocessed to have exclusively continuous attributes, scaled between -1 and 1. Multivalued attributes were binarized.

\setlength{\tabcolsep}{12pt}
\begin{table}[t]
	\centering
	\caption{Datasets}
	\label{tab:data}
	\small
	\begin{tabular}{l|rrrr}
		Dataset				& \# Instances	&\# Attrib	& $\Pr(y=1)$\\\hline
		20 Newsgroups 		& 3,979			& 200		& 0.55	\\
		Adult				& 48,842		& 14  		& 0.24	\\
		Breast Cancer		& 683			& 9 		& 0.35	\\
		Cover Type			& 581,0124  	& 54		& 0.49	\\
		Diabetes			& 99,492		& 127		& 0.11 	\\
		Image Segm.			& 2,310 		& 18		& 0.43	\\
		Mushroom			& 8,124			& 111 		& 0.48	\\
		Splice 				& 3,175 		& 60		& 0.52	
	\end{tabular}
\end{table}




\subsection{Methodology and Approaches}

\subsubsection{Constructing Datasets.} First, we extended each dataset with four artificial binary propensity attributes $x_e^{(i)}\in \{0,1\}$. Therefore, we clustered each dataset into five groups (based on the attribute values) and generated for each group a random distribution between propensity attribute values $\{0,1\}$. Intuitively, this corresponds to a scenario where examples that are in the same cluster have the same prior probability of belonging to the positive class. However, which examples are labeled depend on the propensity attributes. 


Next, the datasets were randomly partitioned into train ($80\%$) and test ($20\%$) sets five times. For each of the five train-test splits, we transformed the data into positive and unlabeled datasets by sampling the examples to be labeled according to the following propensity score:

\begin{align*}
	&e(x_e) = \prod_{i=1}^k \left( p_\text{low}^{1-x_e^{(i)}}\cdot p_\text{high}^{x_e^{(i)}} \right)^\frac{1}{k}
\end{align*}

\noindent This gives propensity scores between $p_\text{low}$ and $p_{high}$, with all artificial propensity attributes $x_e$ attributing equally to it. In our experiments the propensity scores were between 0.2 and 0.8. We generated five of such labelings for each of the five train-test splits and report the average performance over these 25 experiments.

\subsubsection{Approach.}  We compare the classification performance of our EM method under the SAR assumption\footnote{Implementation available on \url{https://dtai.cs.kuleuven.be/software/sar}} against four baselines. First, we assume the data is SCAR and compare against two state-of-the art methods to estimate the class prior: KM2 from \citet{ramaswamy2016mixturepe}\footnote{\url{http://web.eecs.umich.edu/~cscott/code/kernel_MPE.zip}} and TI$c$E from \citet{bekker2018aaai}\footnote{\url{https://dtai.cs.kuleuven.be/software/tice/}} with standard settings. Second, we use the naive baseline which assumes that all unlabeled examples belong to the negative class (denoted Naive). Finally, as an illustrative upper bound on performance, we show results when given fully supervised data (denoted Sup.). All approaches use logistic regression with default parameters from Scikit-Learn\footnote{https://scikit-learn.org/stable/} as the base classifier for the classification model and also for the propensity score model in the SAR setting.





\subsection{Results}

\begin{figure}[!h]
	\begin{center}
		\par\bigskip\bigskip 
		\subfloat[Classification performance ($\text{MSE}_f$)\label{fig:learn_f_mse}]{
			\includegraphics[width=.75\paperwidth, center]{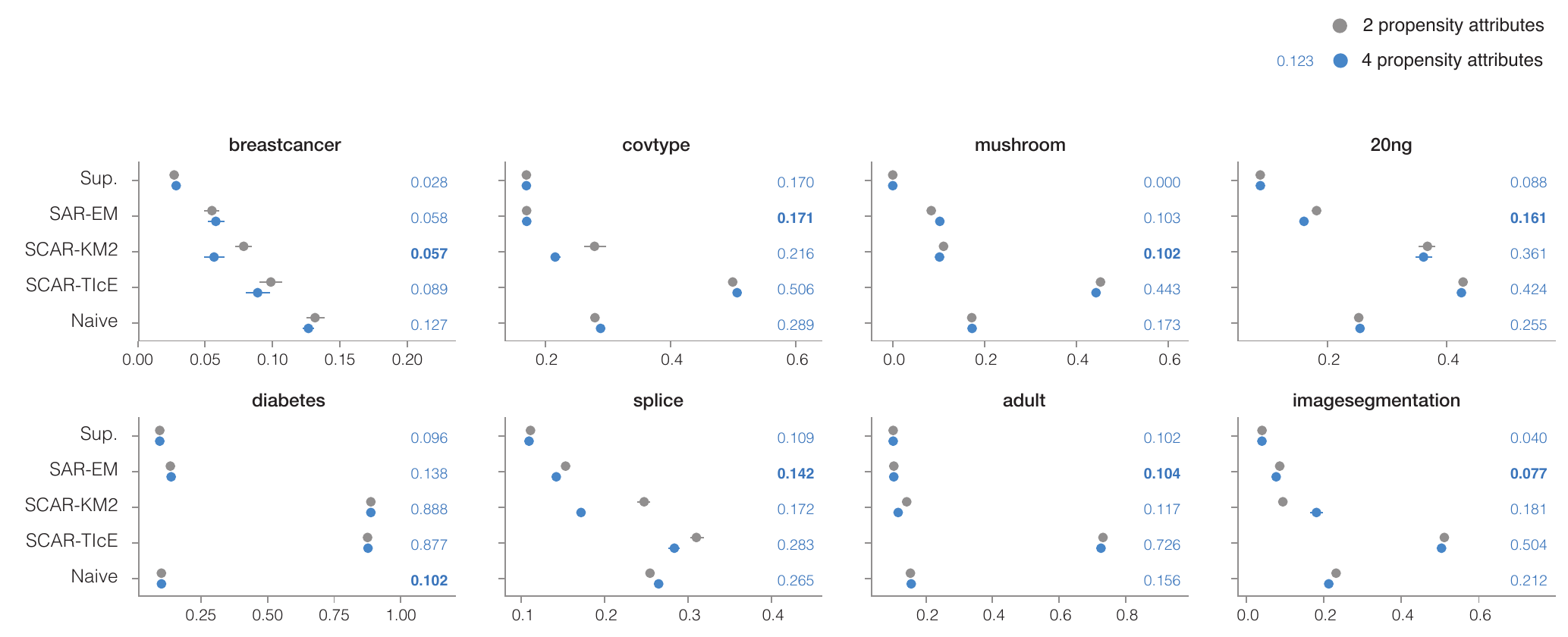}
		}
		\par\bigskip\bigskip 
		
		\subfloat[Classification performance ($\text{ROC-AUC}_f$)\label{fig:learn_f_roc}]{
			\includegraphics[width=.75\paperwidth, center]{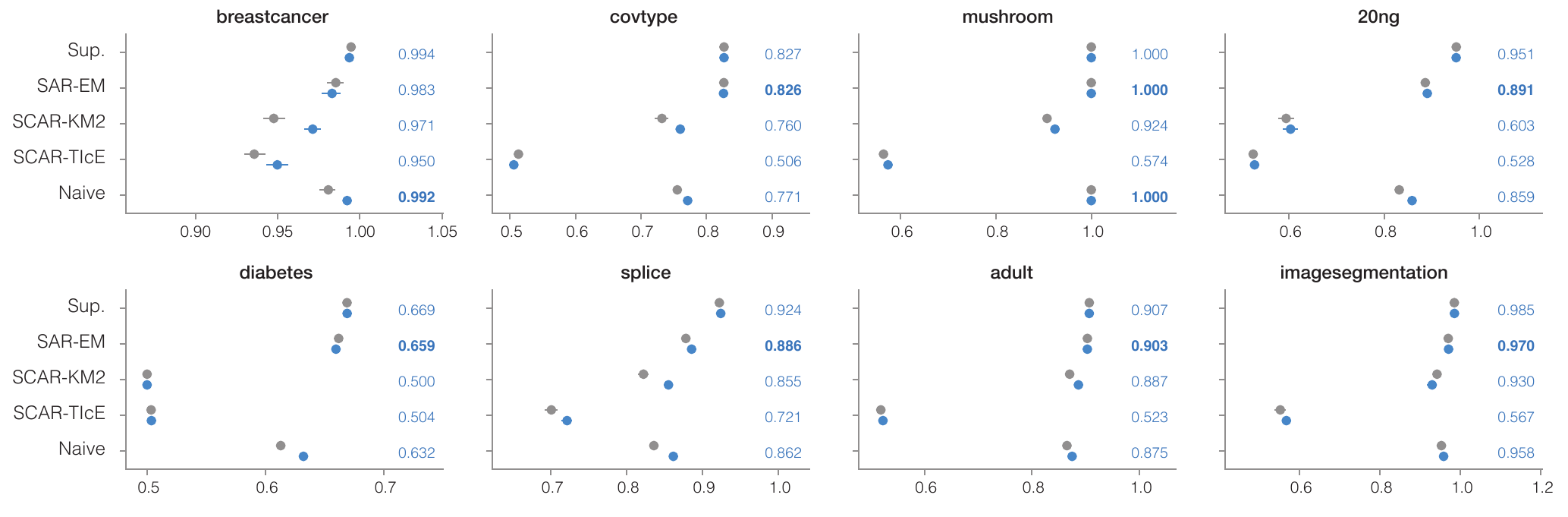}
		}
		
		
		\par\bigskip\bigskip 
		
		\caption{Given SAR data, jointly learning both the unknown propensity scores and the classification model almost always outperforms learning under the SCAR assumption. The dots correspond to the mean performance for respectively two (grey) and four (blue) propensity attributes. The error bars represent a 95\% confidence interval around the mean. The exact performance metric value is given on the right for the setting with four propensity attributes, with the best performing algorithm highlighted in bold (ignoring Supervised).}
		\label{fig:learn_nbf}
	\end{center}
\end{figure}

\textbf{Q1.~} Figure~\ref{fig:learn_nbf} compares SAR-EM to all baselines. Because we are considering models that predict probabilities for binary classification problems, we report two standard metrics. First, we report MSE which measures the quality of the model's probability estimates~\cite{flach2019evaluation}. Second, we report ROC-AUC, which measures predictive performance. When the propensity attributes are known, learning both the propensity score and the classification model from the data outperforms assuming the data is SCAR and learning under that assumption. Based on each method's average ROC-AUC ranks over all eight benchmark datasets, the Friedman test~\cite{demvsar2006statistical} rejects the null-hypothesis that all methods perform the same ($p < 0.001$), regardless of the number of propensity attributes used. Moreover, using the Nemenyi post-hoc test on the ranks, the performance of our SAR-EM method is significantly better ($p < 0.01$) than the naive approach, KM2 and TI$c$E. Note that the naive approach sometimes outperforms the SCAR approaches. This can be explained by the SCAR methods' goal of predicting the correct ratio of the instance space as positive. When it picks the wrong subpsace to get to this ratio, it results in both false positives and false negatives, where the more conservative naive approach would only give the false negatives. 

\FloatBarrier

\begin{figure}[!h]
	\centering
	\includegraphics[width=.75\paperwidth, center]{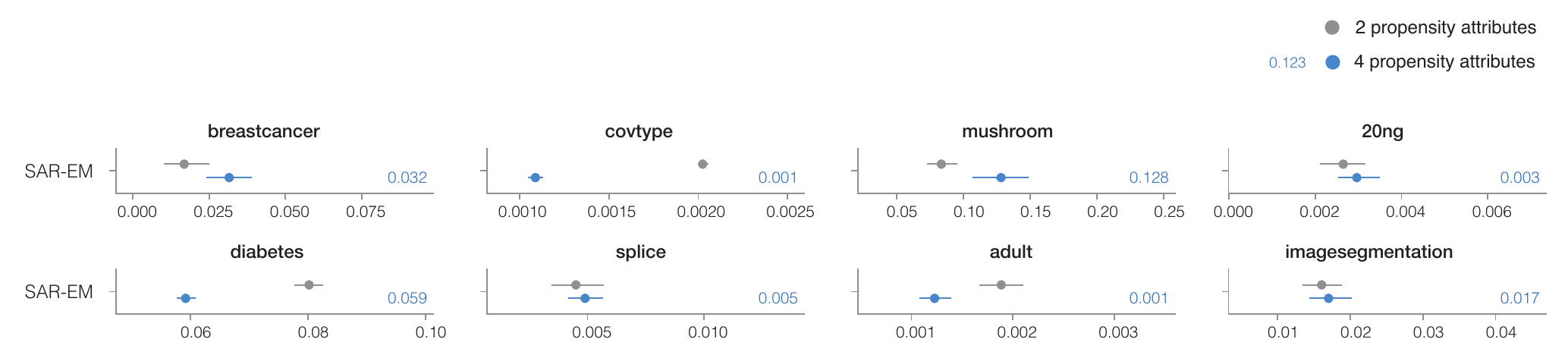}
	\caption{Accuracy of the propensity score estimates ($\text{MSE}_e$).  The dots correspond to the mean performance for respectively two (grey) and four (blue) propensity attributes. The error bars represent a 95\% confidence interval around the mean. The exact performance metric value is given on the right for the setting with four propensity attributes.}
	\label{fig:learn_e}
\end{figure}

\textbf{Q2.~} To evaluate the quality of the learned propensity scores for each example, we report the MSE~\cite{flach2019evaluation}. Except for the mushroom dataset, the EM method always obtains very accurate propensity score estimates with MSEs below 0.1 (Figure~\ref{fig:learn_e}). Furthermore, the MSEs are often very close to zero.   

\textbf{Q3.~} Finally, we observe no correlation between the number of propensity attributes and the MSE and ROC-AUC of the classification model, nor the MSE of the propensity score estimates (Figure~\ref{fig:learn_f_mse}).



\section{Related Work}


PU learning is an active area and for a broad overview see~\cite{bekker2018:survey}. This work focuses on approaches that modify learning methods by exploiting the assumptions about the labeling mechanism (e.g.,~\cite{denis2005learning,elkan2008learning,hsieh2015pu,lee2003learning,Northcutt2017LearningWC,ward2009presence}) for the single training set scenario. 
The key difference is that this paper generalizes past work, which has focused on the SCAR assumption, to the less restrictive SAR setting.  The weighting scheme used in this paper has been used under the SCAR assumption~\cite{steinberg1992estimating,Plessis2015ConvexFF,Kiryo2017PositiveUnlabeledLW}. Furthermore, a special case of this method has been applied matrix completion~\cite{hsieh2015pu}.

Almost all PU learning work that we are aware of focuses on the SCAR setting. One recent exception assumes that the probability of observing a label for a positive example depends on how difficult the example is to label~\cite{he2018idpu}. That is, the more similar a positive example is to a negative one, the less likely it is to be labeled. The difficulty of labeling is defined by the \emph{probabilistic gap} $\Delta\Pr(x) = \Pr(y=1|x)-\Pr(y=0|x)$~\cite{he2018idpu}. Based on properties of the probabilistic gap, it is possible to identify reliable positive and negative examples~\cite{he2018idpu}. Because the probabilistic gap labeling mechanism depends on the attribute values $x$, it is a specific case of SAR. Concretely, it assumes a propensity score that is a non-negative, monotonically decreasing function of the probabilistic gap $\Delta\Pr(x)$.


PU learning is a special case of semi-supervised learning~\citep{chapelle2009semi}. It is also related to one-class learning~\citep{khan_madden_2014}. The work  on dealing with biases in the observed ratings for recommender systems~\cite{Schnabel2016RecommendationsAT} and implicit feedback~\cite{joachims:wsdm17} is closely related to ours. They also make use of propensity scores to cope with the biases. However, there is a crucial difference: they perform inverse propensity weighting, which is not possible in our setting.  In those works, the propensity score for each example is non-zero. In contrast, in PU learning, the propensity score for any negative example is zero: you never observe these labels. Moreover, they assume that examples for the full label space are available (e.g., observe at least one rating of each category for recommender systems) to learn the propensity model, which is not the case for PU learning because we have no known negative examples. This necessitates different ways to learn the propensity scores and weigh the data in our setting. 

\section{Conclusions}
We investigated learning from SAR PU data: positive and unlabeled data with non-uniform labeling mechanisms. We proposed and theoretically analyzed an empirical-risk-minimization based method for weighting PU datasets with the propensity scores to achieve unbiased learning. We explored which assumptions are necessary to learn from SAR PU data generated by an unknown labeling mechanism and proposed a practical EM-based method for this setting. Empirically, for SAR PU data, our proposed propensity weighted method offers superior predictive performance over making the SCAR assumption. Moreover, we are able to accurately estimate each example's propensity score. 


%
%
%

\section*{Acknowledgments}
JB is supported by IWT(SB/141744). PR is supported by by Nano4Sports and Research Foundation - Flanders (G0D8819N). JD is partially supported by KU Leuven Research Fund (C14/17/07, C32/17/036), Research Foundation - Flanders (EOS No. 30992574, G0D8819N) and VLAIO-SBO grant HYMOP (150033).

\bibliography{references}
\bibliographystyle{splncsnat}

\fullpaper{
\clearpage
\appendix
\input{appendix.tex}
}

\end{document}

%% file: appendix.tex
\section{Proof of Proposition 1}
\setcounter{proposition}{0}
\begin{proposition}[Propensity-Weighted Estimator Bound]
		For any predicted classes $\hat \y$ and real classes $\y$ of size $n$, with probability $1-\eta$, the propensity-weighted estimator $\hat R(\hat \y|\s,\e)$ does not differ from the real evaluation measure $R(\hat \y|\y)$ more than
		\begin{align*}
		|\hat R(\hat \y|\e,\s)-R(\hat \y|\y)|\leq \sqrt{\frac{\delta_\text{max}^2\ln\frac{2}{\eta}}{2n}},
		\end{align*}
		with $\delta_\text{max}$ the maximum absolute value of cost function $\delta_y$.
	\end{proposition}

\begin{proof}
All the examples are selected to be labeled independently from each other. Therefore, the weighted costs of the examples are independent random variables. As a result, the Hoeffding inequality can be applied \cite{hoeffding1963probability}:
\begin{align*}
\Pr(|\hat R(\hat \y|\e,\s)-\mathbb{E}[\hat R(\hat \y|\e,\s)]| \geq \epsilon) &\leq 2 \exp\left(\frac{-2n\epsilon^2}{\delta_\text{max}^2}\right)\\
\Leftrightarrow \Pr(|\hat R(\hat \y|\e,\s)-R(\hat \y|\y)| \geq \epsilon) &\leq 2 \exp\left(\frac{-2n\epsilon^2}{\delta_\text{max}^2}\right)
\end{align*}

By setting defining the right-hand side of the inequality to $\eta$, the bound $e$ can be calculated in terms of $\eta$:
\begin{align*}
\eta =  2 \exp\left(\frac{-2n\epsilon^2}{\delta_\text{max}^2}\right)
\end{align*}
\begin{align*}
\epsilon = \sqrt{\frac{\delta_\text{max}^2\ln\frac{2}{\eta}}{2n}}.
\end{align*}
\end{proof}

\section{Proof Proposition 2}
\begin{proposition}[Propensity-Weighted ERM Generalization Error Bound]
	For a finite hypothesis space $\mathcal{H}$, the difference between the propensity-weighted risk of the empirical risk minimizer $\hat \y_{\hat R}$  and its true risk is bounded, with probability $1-\eta$, by:	
	\begin{align*}
	R(\hat \y_{\hat R}|\y) \leq \hat R(\hat \y_{\hat R}|\e,\s) + \sqrt{\frac{\delta_\text{max}^2\ln\frac{|\mathcal{H}|}{\eta}}{2n}}
	\end{align*}
\end{proposition}

\begin{proof}
\begin{align*}
&\Pr\left(\hat R(\hat \y_{\hat R}|\e,\s)-R(\hat \y_{\hat R}|\y) \geq \epsilon\right)\\
&\leq \Pr\left(\max_{\hat \y_i}(\hat R(\hat \y_i|\e,\s)-R(\hat \y_i|\y)) \geq \epsilon\right)\\
&= \Pr\left(\bigvee_{\hat \y_i}(\hat R(\hat \y_i|\e,\s)-R(\hat \y_i|\y)) \geq \epsilon\right)\\
&\#\text{\textit{ Boole's inequality}}\\
&\leq \sum_{i=1}^{|\mathcal{H}|} \Pr\left(\hat R(\hat \y_i|\e,\s)-R(\hat \y_i|\y) \geq \epsilon\right)\\
&\#\text{\textit{ Hoeffding's inequality }}\\
&\leq |\mathcal{H}| \cdot \exp\left(\frac{-2n\epsilon^2}{\delta_\text{max}^2}\right) = \eta\\
&\#\text{\textit{ Solve for }}\epsilon\\
&\epsilon = \sqrt{\frac{\delta_\text{max}^2\ln\frac{|\mathcal{H}|}{\eta}}{2n}}
\end{align*}
\end{proof}

\section{Proof of Proposition 3}

\begin{proposition}[Propensity-Weighted Estimator Bias]
	\begin{align*}
	\text{bias}(\hat R(\hat \y|\hat \e,\s)) = \frac{1}{n}\sum_{i=1}^n y_i(1-\frac{e_i}{\hat e_i})\left(\delta_1(\hat y_i) -\delta_0(\hat y_i)\right)
	\end{align*}
\end{proposition}

\begin{proof}
	\begin{align*}
	\text{bias}(\hat R(\hat \y|\hat \e,\s)) &= R(\hat \y) - \mathbb{E}[\hat R(\hat \y|\hat \e,\s)]
	\end{align*}
	
	\begin{align*}
&\mathbb{E}[\hat R(\hat \y|\hat \e,\s)]\\
&\qquad = \frac{1}{n}\sum_{i=1}^n y_i e_i\left(\frac{1}{\hat e_i}\delta_1(\hat y_i) + (1-\frac{1}{\hat e_i})\delta_0(\hat y_i)\right)\\
&\qquad\qquad\qquad + (1-y_i e_i)\delta_0(\hat y_i)\\
&\qquad = \frac{1}{n}\sum_{i=1}^n y_i \frac{e_i}{\hat e_i}\delta_1(\hat y_i) + (1-y\frac{e_i}{\hat e_i})\delta_0(\hat y_i)
	\end{align*}
	
\begin{align*}
&\text{bias}(\hat R(\hat \y|\hat \e,\s))\\
&\qquad =  \frac{1}{n}\sum_{i=1}^n (y_i-y_i\frac{e_i}{\hat e_i})\delta_1(\hat y_i) + (1-y_i - 1+y\frac{e_i}{\hat e_i})\delta_0(\hat y_i)\\
&\qquad =  \frac{1}{n}\sum_{i=1}^n y_i(1-\frac{e_i}{\hat e_i})\delta_1(\hat y_i) - y_i(1-\frac{e_i}{\hat e_i})\delta_0(\hat y_i)\\
&\qquad =  \frac{1}{n}\sum_{i=1}^n y_i(1-\frac{e_i}{\hat e_i})\left(\delta_1(\hat y_i) -\delta_0(\hat y_i)\right)\\
\end{align*}
	
\end{proof}

\section{Expected Risk Derivation}
The expected risk is defined as

\begin{align*}
\hat R_\text{exp}(\hat \y|\e,\s) = \mathbb{E}_{\y|\e,\s,\hat \y}\left[R(\hat\y|\y)\right]
& = \mathbb{E}_{\y|\e,\s,\hat \y}\left[\frac{1}{n}\sum_{i=1}^n y_i\delta_1(\hat y_i) + (1- y_i)\delta_0(\hat y_i)\right]\\
& = \frac{1}{n}\sum_{i=1}^n \Pr(y_i=1|e_i,s_i,y_i)\delta_1(\hat y_i) \\
&\qquad\qquad + (1- \Pr(y_i=1|e_i,s_i,y_i))\delta_0(\hat y_i).
\end{align*}

With conditional probabilities $\Pr(y_i=1|e_i,s_i,y_i)$:
\begin{align*}
\Pr(y_i=1|e_i,s_i,\hat y_i)
&= s\Pr(y_i=1|e_i,s_i=1,\hat y_i)\\
&\qquad+(1-s)\Pr(y_i=1|e_i,s_i=0,\hat y_i)\\
&= s+(1-s)\frac{\Pr(y_i=1|\hat y_i, e_i)\Pr(s=0|\hat y=1,\hat y_i,e_i)}{\Pr(s=0|\hat y,e_i)}\\
&= s+(1-s)\frac{\hat y_i(1-\Pr(s=1|\hat y=1,\hat y_i,e_i))}{1-\Pr(s=1|\hat y,e_i)}\\
&= s+(1-s)\frac{\hat y_i(1-e_i)}{1-\hat y_i e_i},\\
\end{align*}

this results in 
\begin{align*}
\hat R_\text{exp}(\hat \y|\e,\s)
&= \frac{1}{n}  \sum_{i=1}^n \left(s_i+(1-s_i)\frac{\hat y_i(1-e_i)}{1-\hat y_ie_i}\right)\delta_1(\hat y_i)+(1-s_i)\frac{1-\hat y_i}{1-\hat y_ie_i}\delta_0(\hat y_i).
\end{align*}

\section{Expectation Maximization Derivation}

\subsection{Maximization}
	
	\begin{align*}
	f,e = &\argmax_{f,e}  \sum_{i=1}^n \mathbb{E}_{y_i|x_i,s_i,\hat f,\hat e}\ln \Pr(x_i,s_i,y_i|f,e)\\
	&= \argmax_{f,e}  \sum_{i=1}^n \mathbb{E}_{y_i|x_i,s_i,\hat f,\hat e} \ln\left[ \Pr(x_i)\Pr(y_i|x_i,f)\Pr(s_i|y_i,x_i,e)\right] \\
	&= \argmax_{f,e}  \sum_{i=1}^n \mathbb{E}_{y_i|x_i,s_i,\hat f,\hat e} \ln \left[\Pr(y_i|x_i,f)\Pr(s_i|y_i,x_i,e)\right] ~ \text{\small{\textit{ max not over $\Pr(x_i)$}}}\\
	&= \argmax_{f} \sum_{i=1}^n\mathbb{E}_{y_i|x_i,s_i,\hat f,\hat e} \ln \Pr(y_i|x_i,f)\\
	& \quad + \argmax_{e}\sum_{i=1}^n\mathbb{E}_{y_i|x_i,s_i,\hat f,\hat e} \ln\Pr(s_i|y_i,x_i,e)\\
	&= \argmax_{f} \sum_{i=1}^n \left[\hat y_i \ln \Pr(y_i=1|x_i,f) 
	 + (1-\hat y_i)\ln\Pr(y_i=0|x_i,f) \right],\\
	& \quad \argmax_{e} \sum_{i=1}^n \left[\hat y_i \ln \Pr(s_i|y_i=1,x_i,e) 
	 + (1-\hat y_i)\ln\Pr(s_i|y_i=0,x_i) \right]\\
	 &= \argmax_{f} \sum_{i=1}^n \left[\hat y_i \ln \Pr(y_i=1|x_i,f) 
	 + (1-\hat y_i)\ln\Pr(y_i=0|x_i,f) \right],\\
	 & \quad \argmax_{e} \sum_{i=1}^n \left[\hat y_i \ln \Pr(s_i|y_i=1,x_i,e)  \right]\quad\text{\small{\textit{ max not over $\Pr(s_i|y_i=0,x_i)$}}}\\
	 &= \argmax_{f} \sum_{i=1}^n \hat y_i \ln \Pr(y_i=1|x_i,f) 
	 + (1-\hat y_i)\ln\Pr(y_i=0|x_i,f),\\
	 & \quad \argmax_{e} \sum_{i=1}^n \hat y_i \big[s_i \ln \Pr(s_i=1|y_i=1,x_i,e)\\
	 &  \qquad \qquad \qquad \quad+ (1-s_i) \ln (1-\Pr(s_i=1|y_i=1,x_i,e))  \big]\\
	\end{align*}

To avoid a convoluted derivation, we use $\hat y_i$ as a shorthand for $\Pr(y_i=1|s_i,x_i,\hat f,\hat e)$.
\subsection{Expectation}

\begin{align*}
	\Pr(y=1|s,x,f,e) &=  s \Pr(y=1|s=1,x) + (1-s)\Pr(y=1|s=0,x)\\
	&= s  + (1-s)\frac{\Pr(y=1|x)\Pr(s=0|y=1,x)}{\Pr(s=0|x)}\\
	&= s  + (1-s)\frac{\Pr(y=1|x)\left(1-\Pr(s=1|y=1,x)\right)}{1-\Pr(s=1|x)}\\
	&= s  + (1-s)\frac{\Pr(y=1|x)\left(1-\Pr(s=1|y=1,x)\right)}{1-\Pr(y=1|x)\Pr(s=1|y=1,x)},\\
\end{align*}
where the first step follows from the definition of PU data $s=1\rightarrow y=1$ and Bayes' rule.